\documentclass{article}
\usepackage{spconf,amsmath,graphicx,hyperref}

\usepackage{microtype}

\usepackage{booktabs}  
\usepackage{arydshln}

\usepackage{multirow}
\usepackage{float}
\usepackage{mismath}
\usepackage[many]{tcolorbox}
\usepackage{amssymb}
\usepackage{mathtools}
\usepackage{amsthm}
\usepackage{pifont}
\usepackage{enumitem}
\usepackage{CJKutf8}

\makeatletter
\def\eg{\textit{e.g.}} 
\def\ie{\textit{i.e.}}

\makeatother

\newcommand{\keypoint}[1]{\noindent \textbf{#1}:}

\usepackage{subcaption}

\usepackage{algorithm}
\usepackage[noend]{algpseudocode}

\usepackage[english]{babel} 

\usepackage{tikz}
\usetikzlibrary{decorations.pathreplacing,calc}

\tikzset{brace/.style={decorate, decoration={brace}},
 brace mirrored/.style={decorate, decoration={brace,mirror}},
}

\newcounter{brace}
\newcounter{arrow}
\title{Data-driven Clustering and Merging of Adapters \\for On-device Large Language Models}
\name{Ondrej Bohdal\textsuperscript{1}, Taha Ceritli\textsuperscript{1}, Mete Ozay\textsuperscript{1}, Jijoong Moon\textsuperscript{2},}
\nametwo{Kyeng-Hun Lee\textsuperscript{2}, Hyeonmok Ko\textsuperscript{2}, Umberto Michieli\textsuperscript{1}}
\address{\textsuperscript{1}Samsung R\&D Institute UK, United Kingdom \ \textsuperscript{2}Samsung Research, South Korea}

\begin{document}
\maketitle
\begin{abstract}
On-device large language models commonly employ task-specific adapters (\eg, LoRAs) to deliver strong performance on downstream tasks. While storing all available adapters is impractical due to memory constraints, mobile devices typically have sufficient capacity to store a limited number of these parameters. This raises a critical challenge: how to select representative adapters that generalize well across multiple tasks—a problem that remains unexplored in existing literature.
We propose a novel method $D^2C$ for adapter clustering that leverages minimal task-specific examples (\eg, 10 per task) and employs an iterative optimization process to refine cluster assignments. The adapters within each cluster are merged, creating multi-task adapters deployable on resource-constrained devices. 
Experimental results demonstrate that our method effectively boosts performance for considered storage budgets.
\end{abstract}
\section{Introduction}
\label{sec:introduction}
Large language models (LLMs) have emerged as versatile foundations for diverse text generation tasks including summarization \cite{liu2023learning}, question answering \cite{sticha2024qa}, and multilingual translation \cite{zhu2023multilingual}, as comprehensively surveyed in \cite{zhao2023survey}. Parameter-efficient fine-tuning (PEFT) methods such as Low-Rank Adaptation (LoRA) \cite{hu2021lora} enable effective specialization of these foundation models for downstream tasks while preserving base model integrity \cite{han2024parameter}.

The deployment of smaller LLMs (\eg, of size 1B to 5B) on mobile devices presents unique advantages in privacy preservation and operational cost reduction by eliminating cloud dependencies. 
However, smaller LLMs have limited zero-shot abilities; hence, current practice employs multiple LoRA adapters to equip these smaller models with task-specific capabilities \cite{gunter2024apple}. Additionally, this paradigm allows for easy addition of new functionalities by modifying the adapters rather than the whole model.

\begin{figure}[t]
\vskip 0.2in
\begin{center}
\centerline{\includegraphics[width=0.8\columnwidth]{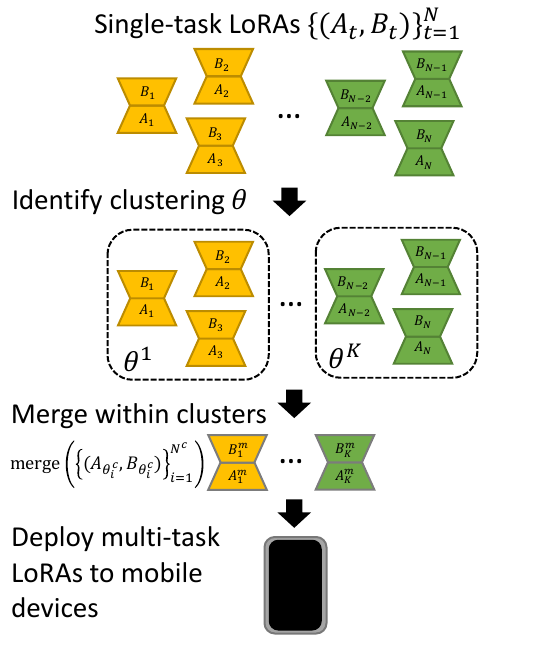}}
\caption{\textbf{Overall pipeline.} Single-task LoRAs are first clustered and then merged within the clusters. The resulting multi-task LoRAs are deployed to mobile devices under strict storage constraints.
}
\label{fig:overview}
\end{center}
\vskip -0.2in
\end{figure}

There are numerous use cases for LLMs, but there is only a limited amount of storage available on a mobile device. As a result, it is not feasible to store a separate LoRA adapter for each specific task. At the same time, it is common to train single-task LoRAs, for example, due to challenges in data sharing, distribution of expertise, or simply the inconvenience of having to retrain on all data when new functionalities are added. A solution is to merge already-trained LoRA adapters \cite{yang2024model}. However, how to merge LoRAs remains an unexplored problem, especially when we can store a few LoRAs, rather than just one. 

In this work, we aim to automatically find how to cluster the LoRAs and then merge them within the corresponding groups using existing merging techniques. As part of our algorithm, we assume access to a few examples from each task and iteratively find a good way to cluster the LoRAs. Once the clustering is finished, we merge the LoRAs and deploy them to mobile devices. We provide an illustration in Figure~\ref{fig:overview}.

Our key contribution is a simple yet well-performing algorithm $D^2C$ for identifying clusters of LoRAs so that we can merge them and obtain a desired number of multi-task LoRAs.

\section{Related Work}
\label{sec:related}

\textbf{Parameter-efficient Fine-tuning (PEFT)} approaches adapt models efficiently by training relatively few parameters, making them especially suitable for fine-tuning LLMs \cite{han2024parameter}. LoRA \cite{hu2021lora} introduces low-rank matrices into selected layers and trains only these additional parameters. It has small additional storage requirements because of its compact size, making it suitable for deployment to edge devices \cite{gunter2024apple}. LoRA has been further improved in its numerous extensions, such as Delta-LoRA \cite{zi2023delta}, DoRA \cite{liu2024dora}, and RAC-LoRA \cite{malinovsky2024randomized}.

\keypoint{Model Merging} Task-specific models can be combined via model merging into one model that can do multi-tasking. Task Arithmetic \cite{wortsman2022model,ilharco2022editing} combines the weights of multiple models as a weighted average and can be seen as the simplest option. TIES \cite{yadav2024ties} and DARE \cite{yu2024language} are examples of popular, more advanced merging methods. TIES begins by resetting the values of parameters that have changed only a little, then it elects the sign if there are conflicts, and finally it merges only sign-aligned parameters. DARE drops part of the weight changes and rescales the remaining ones. Other methods \cite{huang2023lorahub,hammoud2024model,shenaj2024lora} use data to improve merging, but the data is not used to determine the clusters. Recent advanced use cases of model merging include compositional multi-tasking \cite{bohdal2025compositional} and continual merging \cite{shenaj2025continual}. Further, efficiency-performance trade-off of model merging has been studied in \cite{ceritli2025hydraopt}.

\section{Method}
We are provided with a budget of $K$ LoRAs that we can store on a mobile device, but there are $N>K$ single-task LoRAs, each supporting a different functionality. We assume training new multi-task LoRAs is not practical, \eg, due to issues with data sharing, expertise of the individual LoRA developers, or simply to avoid retraining the LoRAs. As we can store only $K<N$ LoRAs, we need to find a way to group the LoRAs into $K$ clusters via a partitioning map $\theta$ so that the LoRAs within each cluster can then be merged using existing techniques such as TIES \cite{yadav2024ties}. Map $\theta$ returns original LoRA indices for each cluster, \ie,~$\theta^c_i$ returns the original index of the $i$-th LoRA in cluster $c$.
Common partitioning options include random or K-Means clustering of LoRAs. We improve upon these data-free options by assuming access to a few examples for each task (\eg, $n=10$ examples). These can be easily hand-crafted if no access is granted to the original training data, otherwise, random samples from training set can be used.

At the very beginning, the LoRAs are randomly partitioned into $K$ clusters.
We introduce an iterative algorithm $D^2C$ where, in each iteration, we randomly select two clusters $c, c^\prime$ and a LoRA $(A_t, B_t)$ (with associated task $t$) from cluster $c$.
The algorithm computes a merged LoRA from the set of $N^c$ LoRAs that belong to cluster $c$, \ie, $\{(A_{\theta^c_i}, B_{\theta^c_i})\}_{i=1}^{N^c}$, via a $\text{merge}$ operation (\eg, TIES). The merged LoRA is evaluated on data from task $t$, $\mathcal{D}_t \subset \mathcal{D}$ ($|\mathcal{D}_t|=n$, $|\mathcal{D}|=N \times n$ for $N$ tasks), to compute cross-entropy loss $\ell$. Then, the LoRA $(A_t, B_t)$ is moved to cluster $c^\prime$ and the loss $\ell^\prime$ is computed over $\mathcal{D}_t$. If the loss is not reduced, we return LoRA $(A_t, B_t)$ to cluster $c$.
We run the algorithm for a given number of iterations $T$. After finishing, we take the final partition map of LoRAs into $K$ clusters and merge the LoRAs within each cluster to obtain $K$ multi-tasking LoRAs. 
Algorithm~\ref{alg:algo} shows a more detailed description of our $D^2C$ approach.

\label{sec:method}
\begin{algorithm}[!htb]
  \caption{Data-Driven Clustering ($D^2C$)}
  \label{alg:algo}
    \begin{algorithmic}[1]
    \Require $K$ clusters, $N$ LoRAs $\{(A_t, B_t)\}_{t=1}^N$, dataset $\mathcal{D}$ with $n$ examples per task, number of iterations $T$
    \State{Randomly initialize LoRA partition map $\theta$} 
    \For {$T$ iterations}
        \State{Sample two cluster indices $c$ and $c^\prime$ }
        \State{Sample LoRA $(A_t, B_t)$ with associated task $t$ from cluster $c$}
        \State{Compute loss $\ell$ on $\mathcal{D}_t$} of $\text{merge}\left(\{(A_{\theta^c_i}, B_{\theta^c_i})\}_{i=1}^{N^c}\right)$ 
        \State{Update $\theta$: Add LoRA $(A_t, B_t)$ to cluster $c^\prime$ and remove it from cluster $c$}
        \State{Compute loss $\ell^\prime$ on $\mathcal{D}_t$ of $\text{merge}\left(\{(A_{\theta^{c^\prime}_i}, B_{\theta^{c^\prime}_i})\}_{i=1}^{N^{c^\prime}}\right)$}
        \If{$\ell \leq \ell^\prime$}
            \State{Update $\theta$: Bring LoRA $(A_t, B_t)$ back to cluster $c$ and remove it from $c^\prime$}
        \EndIf
    \EndFor
    \State {\textbf{Return} final partition map $\theta$ so that LoRAs can be merged within each cluster}
    \end{algorithmic}
\end{algorithm}

\section{Experiments}
\label{sec:experiments}

\begin{table*}[h]
\setlength{\tabcolsep}{9pt}
\begin{center}
{
\begin{tabular}{lcccc}
\toprule
 &  & & \textbf{Avg. Score} ($\uparrow$, \%) & \\
 \cline{3-5}
\textbf{Method} & \textbf{Storage ($\downarrow$, \%)} & \textbf{Llama 3.2 3B} & \textbf{Qwen 2.5 1.5B} & \textbf{StableLM 2 1.6B} \\
\midrule
Zero-shot & \phantom{00}0.0 & 14.7 $\pm$ 0.0 & 15.2 $\pm$ 0.0 & \phantom{0}6.5 $\pm$ 0.0 \\
Separate single-task LoRAs & 100.0 & 32.9 $\pm$ 0.0 & 31.7 $\pm$ 0.0 & 29.0 $\pm$ 0.0 \\
\hdashline
Random clustering & \phantom{0}12.5 & 21.5 $\pm$ 1.0 & 22.5 $\pm$ 1.3 & 15.2 $\pm$ 0.3 \\
K-Means clustering & \phantom{0}12.5 & 20.4 $\pm$ 0.1 & 21.9 $\pm$ 0.1 & 15.5 $\pm$ 0.1  \\
K-Means w/ SVD clustering & \phantom{0}12.5 & 22.3 $\pm$ 1.2 & 22.4 $\pm$ 0.9 & 16.1 $\pm$ 0.2 \\
$D^2C$ (ours) & \phantom{0}12.5 & 26.0 $\pm$ 0.7 & 24.6 $\pm$ 1.1 & 17.3 $\pm$ 0.3 \\
\bottomrule
\end{tabular}
}
\end{center}
\caption{\textbf{
Main aggregate results across various models.} Our method outperforms competing clustering approaches and obtains an excellent average performance while using only 12.5\% of the storage needed for storing each LoRA separately.}
\label{tab:main}
\end{table*}

\subsection{Setup}
\label{sec:setup}

\keypoint{Tasks} We perform experiments across 40 text-generation tasks, spanning 5 problem types and 8 languages. The problem types include Grammar Error Correction, Smart Reply, Summarization, Tone Adjustment (rewriting), and Question Answering, representing a variety of practical, real-world use cases for on-device text generation. The languages considered are English, Spanish, French, German, Italian, Chinese (simplified), Korean, and Japanese.

\keypoint{Datasets} We use Persona-Chat Synthetic \cite{jandaghi2023faithful} for Smart Reply, SAMSum \cite{gliwa2019samsum} for Summarization, Sound Natural \cite{einolghozati2020sound} rephrased using a public fine-tuned model \cite{utsav2023tone} model for Tone Adjustment, and SQuAD \cite{rajpurkar-etal-2016-squad} for Question Answering. These datasets were collected in English, so we translated them into the other languages. We use OPUS-MT \cite{TiedemannThottingal:EAMT2020} for translation to Spanish, French, German, and Italian, and M2M100 \cite{fan2021beyond} for Chinese, Korean, and Japanese. For Grammar Error Correction we use original datasets because translation typically fixes grammar errors \cite{luhtaru2024no}: Write \& Improve for English \cite{bryant2019bea}, Merlin \cite{boyd2014merlin} for Italian, ECSpell \cite{lv2023general} for Chinese, and GitHub Typo Corpus \cite{hagiwara-mita-2020-github} for the other languages. Each dataset is split into training, validation, and test data, as in \cite{ceritli2025hydraopt}. Validation data were used for hyperparameter selection.

\keypoint{Evaluation Metrics} We use metrics commonly used in literature for the specific problems, namely F-05 score \cite{bryant2017automatic} for Grammar Error Correction (and ChERRANT \cite{zhang-etal-2022-mucgec} for Chinese), Weighted ROUGE for Smart Reply, ROUGE-L for Summarization and Tone Adjustment, and F-1 score for Question Answering. We also report storage required, computed as the percentage of storing all LoRAs.

\keypoint{Model} We use Llama 3.2 3B \cite{dubey2024llama}, Qwen 2.5 1.5B \cite{qwen2.5} and StableLM 2 1.6B \cite{bellagente2024stable} instruction-tuned models, as representative examples of models suitable for on-device deployment. The Llama model is used for the further analyses.

\keypoint{Baselines} We compare with various options for clustering, namely 1) assigning adapters to clusters randomly, 2) applying K-Means clustering directly on flattened and concatenated LoRA $A, B$ matrices, 3) applying K-Means clustering on cosine similarities of SVD features extracted from the LoRAs \cite{ostapenko2024towards}. We also report zero-shot performance and the performance of separate single-task LoRAs as target accuracy. We use a prompt specifying the task, \eg~``\textit{Remove all grammatical errors from this text}''.

\keypoint{Hyperparameters} We use LoRAs with rank 32, $\alpha=128$, dropout 0.05, trained with AdamW optimizer with $5\times 10^{-5}$ learning rate and minibatch size 3. As the merging technique, we use TIES \cite{yadav2024ties} with unary weights and 0.5 density factor. Unless otherwise stated, we use 5 clusters and 200 iterations for our algorithm. We use 10 examples per task, randomly selected from the training sets.

\subsection{Main Results}
\label{sec:main_results}

We report the main aggregated results in Table~\ref{tab:main}, reporting average and standard deviation across three repetitions. The target is to store 5 LoRAs on the device instead of 40, equivalent to using 12.5\% of the storage needed for storing all LoRAs. K-Means-based solutions obtain similar average performance to random clustering, and our solution significantly outperforms random clustering by $\sim$2.1--4.5\%. The performance is strong also in the broader context, \eg~with an average performance of 26.0\% compared to the single-task LoRAs upper bound of 32.9\% and zero-shot performance of 14.7\% for Llama 3.2 3B. Note that merging all LoRAs into one would lead to 18.7\% score in this case.

\subsection{Analyses}
\label{sec:analyses}

\begin{figure}[t]
\vskip 0.2in
\begin{center}
\centerline{\includegraphics[width=\columnwidth]{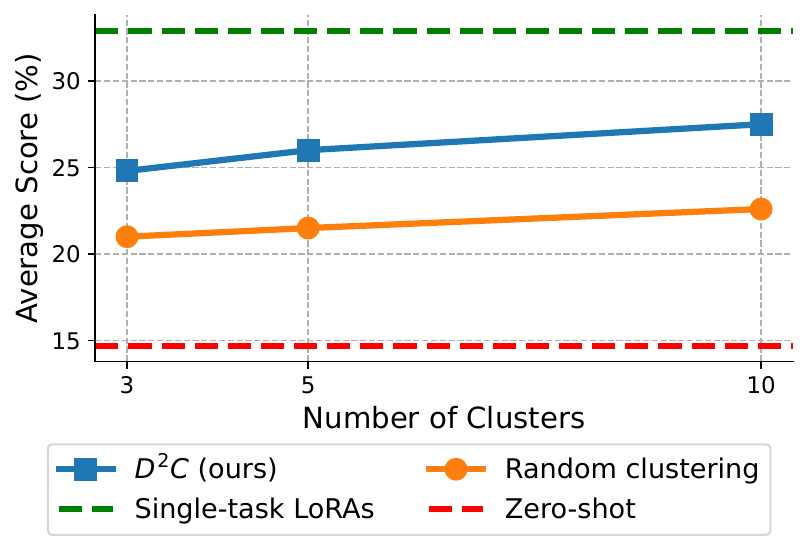}}
\caption{\textbf{Variable number of clusters.} Our solution consistently improves performance and obtains stronger performance with more clusters. If each LoRA was in a separate cluster, there would be 40 clusters.}
\label{fig:num_clusters_analysis}
\end{center}
\vskip -0.2in
\end{figure}

\keypoint{Number of Clusters}
The number of LoRAs that can fit on device storage may vary, so we study how our solution performs for diverse numbers of clusters. The results in Figure~\ref{fig:num_clusters_analysis} show our solution consistently works well for various numbers of clusters, with stronger performance for more clusters, as expected.

\begin{figure}[t]
\vskip 0.2in
\begin{center}
\centerline{\includegraphics[width=\columnwidth]{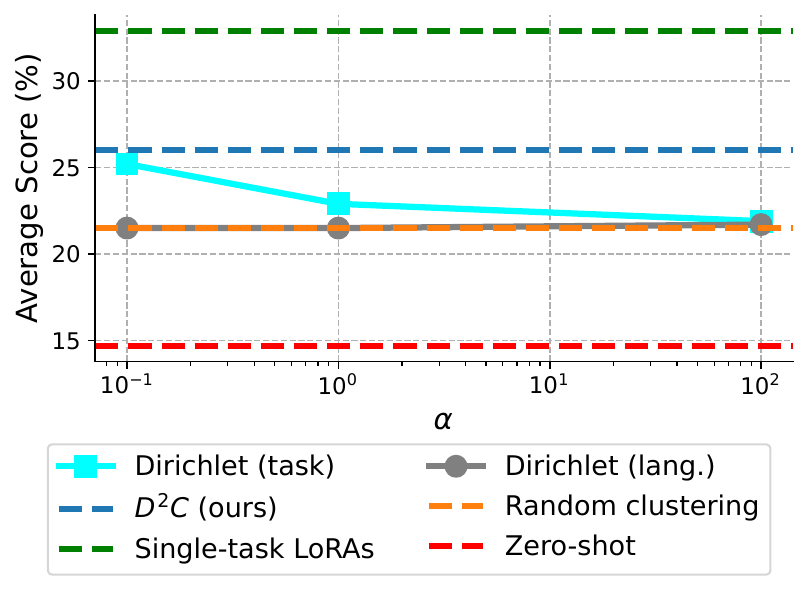}}
\caption{\textbf{Clusters with variable homogeneity.} Smaller $\alpha$ means each cluster is more homogeneous in terms of tasks / languages assigned to it.
The results confirm the findings of our algorithm that it is better to group by task. In contrast, grouping by language does not work well. 
}
\label{fig:dirichlet_analysis}
\end{center}
\vskip -0.2in
\end{figure}

\keypoint{Identified Clusters}
The identified clustering is done predominantly by task. For example, a cluster may only include summarization LoRAs. We investigate further by studying the performance of clusters obtained by Dirichlet sampling, as in \cite{hsu2019measuring}, in terms of tasks and languages, with various levels of temperature $\alpha$, leading to various levels of homogeneity within a cluster. The results in Figure~\ref{fig:dirichlet_analysis} confirm that sampling more homogeneous clusters in terms of the task, \textit{Dirichlet (task)}, leads to stronger performance. In contrast, sampling in terms of language, \textit{Dirichlet (lang.)}, does not bring benefits.

\begin{figure}[t]
\vskip 0.2in
\begin{center}
\centerline{\includegraphics[width=\columnwidth]{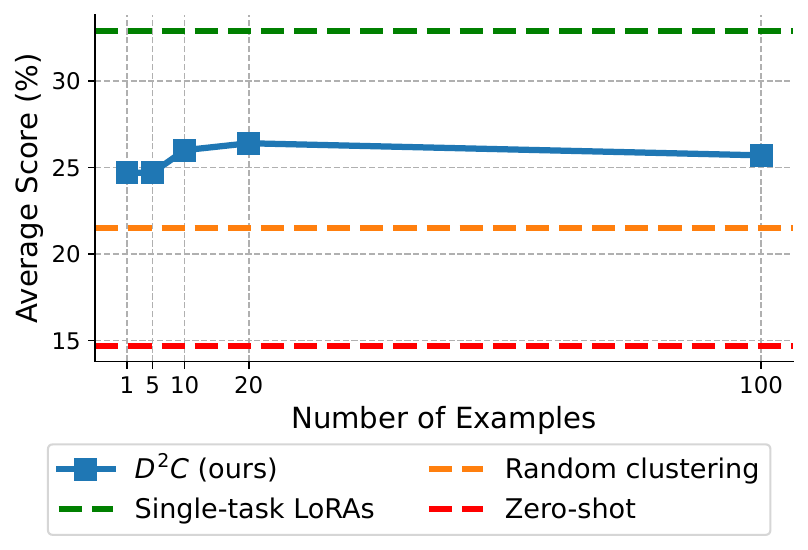}}
\caption{\textbf{Variable number of examples used by our algorithm.} The performance does not vary significantly with the number of examples.}
\label{fig:num_examples_analysis}
\end{center}
\vskip -0.2in
\end{figure}

\keypoint{Number of Examples} Figure~\ref{fig:num_examples_analysis} suggests our algorithm is not particularly sensitive to the number of examples used.

\keypoint{Alternative Merging Strategy} Our method is agnostic to the specific merging technique, and we show it works also with \eg~linear merging instead of TIES. Under this new setting, random clustering achieved 22.1\%, while our solution achieved 24.8\% on average. These results confirm our method maintains its effectiveness across different merging strategies.

\keypoint{Efficiency} With Llama 3.2 3B, the full clustering procedure takes approximately 1.3 hours on a single GPU. This is a modest cost for an offline step, especially considering the resulting performance improvement.

\section{Conclusion}
\label{sec:conclusion}
We have introduced an algorithm $D^2C$ for data-driven clustering and merging of LoRA adapters. The algorithm is simple, uses only a few examples for each task, and obtains strong performance. Using our algorithm, we can automatically find good clusters of single-task LoRAs before merging and so bring strong performance to users while using only a limited amount of storage.

\bibliographystyle{IEEEbib}
{\small\bibliography{custom}}
\end{document}